\pdfoutput=1 
\documentclass[acmsmall]{acmart}

\usepackage{xcolor}
\usepackage{soul}
\usepackage{lscape}
\usepackage{rotating}
\usepackage{array}

\usepackage{pifont}
\usepackage{multirow}
\usepackage{amsmath}
\newcolumntype{P}[1]{>{\centering\arraybackslash}m{#1}}
\newcolumntype{N}{@{}m{0pt}@{}}
\AtBeginDocument{%
  \providecommand\BibTeX{{%
    \normalfont B\kern-0.5em{\scshape i\kern-0.25em b}\kern-0.8em\TeX}}}

\setcopyright{acmcopyright}
\copyrightyear{2021}
\acmYear{2021}
\acmDOI{10.1145/1122445.1122456}

\begin{document}

\title[Counteracting Dark Web Text-Based CAPTCHA with Generative Adversarial Learning] {Counteracting Dark Web Text-Based CAPTCHA with Generative Adversarial Learning for Proactive Cyber Threat Intelligence}

\author{Ning Zhang}
\email{zhangning@email.arizona.edu}
\authornote{Corresponding authors}
\orcid{0000-0003-3799-7337}
\affiliation{%
  \institution{University of Arizona}
  \streetaddress{1130 E. Helen St.}
  \city{Tucson}
  \state{Arizona}
  \country{USA}
  \postcode{85712}
}

\author{Mohammadreza Ebrahimi}
\email{ebrahimim@usf.edu}
\orcid{0000-0003-1367-3338}
\authornotemark[1]
\affiliation{%
  \institution{University of South Florida}
  \streetaddress{4202 E. Fowler Avenue}
  \city{Tampa}
  \state{Florida}
  \country{USA}}

\author{Weifeng Li}
\email{weifeng.li@uga.edu}
\affiliation{%
  \institution{University of Georgia}
  \streetaddress{630 S. Lumpkin St.}
  \city{Athens}
  \country{USA}
}

\author{Hsinchun Chen}
\email{hsinchun@arizona.edu}
\affiliation{%
 \institution{University of Arizona}
 \streetaddress{1130 E. Helen St.}
 \city{Tucson}
 \state{Arizona}
 \country{USA}}

\renewcommand{\shortauthors}{Zhang and Ebrahimi, et al.}

\begin{abstract}
 Automated monitoring of dark web (DW) platforms on a large scale is the first step toward developing proactive Cyber Threat Intelligence (CTI). While there are efficient methods for collecting data from the surface web, large-scale dark web data collection is often hindered by anti-crawling measures. In particular, text-based CAPTCHA serves as the most prevalent and prohibiting type of these measures in the dark web. Text-based CAPTCHA identifies and blocks automated crawlers by forcing the user to enter a combination of hard-to-recognize alphanumeric characters. In the dark web, CAPTCHA images are meticulously designed with additional background noise and variable character length to prevent automated CAPTCHA breaking. Existing automated CAPTCHA breaking methods have difficulties in overcoming these dark web challenges. As such, solving dark web text-based CAPTCHA has been relying heavily on human involvement, which is labor-intensive and time-consuming. In this study, we propose a novel framework for automated breaking of dark web CAPTCHA to facilitate dark web data collection. This framework encompasses a novel generative method to recognize dark web text-based CAPTCHA with noisy background and variable character length. To eliminate the need for human involvement, the proposed framework utilizes Generative Adversarial Network (GAN) to counteract dark web background noise and leverages an enhanced character segmentation algorithm to handle CAPTCHA images with variable character length. Our proposed framework, DW-GAN, was systematically evaluated on multiple dark web CAPTCHA testbeds. DW-GAN significantly outperformed the state-of-the-art benchmark methods on all datasets, achieving over 94.4\% success rate on a carefully collected real-world dark web dataset. We further conducted a case study on an emergent Dark Net Marketplace (DNM) to demonstrate that DW-GAN eliminated human involvement by automatically solving CAPTCHA challenges with no more than 3 attempts. Our research enables the CTI community to develop advanced, large-scale dark web monitoring. We make DW-GAN code available to the community as an open-source tool in GitHub.
\end{abstract}

\ccsdesc[500]{Security and privacy}
\ccsdesc[100]{Information systems, Web searching and information discovery}

\keywords{automated CAPTCHA breaking, dark web, generative adversarial networks}

\maketitle

\section{Introduction}

Illicit platforms have become increasingly common on the dark web \cite{chen_dark_2012}. Examples include Dark Net Markets (DNMs) and carding shops enabling hackers to exchange stolen data, hacking tools, and other illegal products. These malicious products could be detrimental to cyber defense and cause significant loss and damage on cyber infrastructure. To address this issue, proactive Cyber Threat Intelligence (CTI) aims at monitoring the dark web to inform cybersecurity decision making \cite{sapienza2018discover, ebrahimi2018detecting,wen2021key,liu2020identifying, ebrahimi2020detecting, ebrahimi_cross-lingual_nodate}. For instance, cybersecurity firms such as FireEye identify threats to customer assets from the dark web \cite{heires2016terror, Li10.1145/2676869}. Also, with the prevalence of financial data breaches in carding shops and DNMs, monitoring the dark web helps financial institutions alert their customers for potential risks \cite{website}. While developing proactive CTI necessitates automated web crawling for dark web data collection, the dark web extensively employs anti-crawling measures to prevent automated data collection. Text-based CAPTCHA is often known as the most common and difficult anti-crawling measure to counteract in the dark web \cite{weng2019towards, du_identifying_2018}. CAPTCHA stands for Completely Automated Public Turing Test to tell Computers and Human Apart. Text-based CAPTCHA identifies and blocks web crawlers by presenting a heavily obfuscated image of characters and testing the user’s ability to identify the characters shown in the CAPTCHA image. The obfuscation generally involves distorting characters in terms of their font, color, and rotation.

Text-based CAPTCHA can significantly hamper the automation of large-scale dark web collection. When navigating through dark web platforms, crawlers are frequently disrupted by text-based CAPTCHA challenges, which then requires human involvement and is thus detrimental to automated large-scale dark web collection. While recent machine learning (ML) methods have been developed and shown promising results in automated CAPTCHA breaking \cite{ferreira2019breaking, ye2018yet, hussain2017segmentation, george2017generative, gao2013robustness}, dark web platforms increase the difficulty of algorithmic CAPTCHA breaking. Text-based CAPTCHA images in the dark web are different from those on regular platforms in two ways. First, CAPTCHA backgrounds are made particularly noisy by adding distracting background consisting of colorful curves and dots. Such noisy backgrounds are challenging for ML-based methods because they need to learn and distinguish a large number of background patterns, in addition to recognizing the characters. Second, while the character length of text-based CAPTCHA images is a key configuration parameter for training ML-based CAPTCHA breaking methods, dark web platforms rarely use CAPTCHA images with a fixed character length. Pre-trained ML methods often have difficulty in breaking CAPTCHA images with different lengths from what they are trained on. In addition to these challenges, there is a lack of human-labeled dark web-specific text-based CAPTCHA datasets for training ML-based CAPTCHA breaking methods. As such, existing CAPTCHA breaking methods often face trouble with effectively facilitating the automated collection of dark web content for CTI.

Motivated by these challenges, our study contributes to the cybersecurity literature by proposing a novel CAPTCHA breaking framework that leverages the state-of-the-art deep learning techniques for breaking dark web CAPTCHA. The proposed framework, named DW-GAN, comprises three sequential components. First, the automated background removal component seeks to remove the dark web noisy background to improve CAPTCHA breaking performance. To this end, we propose a Generative Adversarial Network (GAN) model, CAPTCHA GAN, that learns to generate CAPTCHA images with relatively clean background from the original patterns with noisy backgrounds. Instead of using millions of labeled CAPTCHA images for training, our CAPTCHA GAN can generate training data to address the lack of labeled training data. Second, the character segmentation component addresses the challenge of variable character length by extracting image segments, each of which contains one single character. Additionally, we extend the core image segmentation technique with a region enlargement procedure for achieving better segmentation. Lastly, the character recognition component detects the character in each CAPTCHA image segment. We utilize the state-of-the-art Convolutional Neural Network (CNN) to recognize segmented characters.

Our proposed framework was rigorously evaluated on a research testbed comprising various CAPTCHA images from the dark web. Our evaluation experiments show that the proposed framework consistently outperformed the state-of-the-art baseline CAPTCHA-breaking methods. In particular, our ablation analysis show that our method was able to effectively remove the colorful curves and dots in the background. Moreover, our proposed CAPTCHA segmentation was able to further improve the success rate of breaking CAPTCHA with variable character length over the state-of-the-art baseline techniques. We demonstrate the applicability of our proposed framework through a case study on dark web data collection, where we incorporated our framework into a dark web crawler. Equipped with DW-GAN, the crawler successfully collected a medium-sized Dark Net Marketplace within almost 5 hours without any human involvement, where DW-GAN was able to break CAPTCHA images with no more than three attempts.

The remainder of the paper is organized as follows. Section 2 presents a review of related literature to motivate our research and provide a technical background of CAPTCH-breaking. Section 3 details our proposed design of the DW-GAN framework. Section 4 systematically examines the effectiveness of our proposed framework and its major novelties in comparison with the state-of-the-art benchmark techniques. Section 5 demonstrates the applicability of our proposed framework through a case study on dark web collection. Lastly, Section 6 summarizes our contribution and discusses our future directions.

\section{Literature Review}
Based on our research objectives, we review two major areas of research. First, we examine dark web data collection for proactive CTI as the major application domain of our study and investigate CAPTCHA as a major challenge hampering dark web data collection. Second, we review the state-of-the-art text-based CAPTCHA breaking methods in prior research. In particular, we explore image preprocessing and background denoising techniques as necessary steps for breaking the dark web CAPTCHA with complex backgrounds. Subsequently, we examine character segmentation as a vital step to breaking CAPTCHA with variable character length. Lastly, we investigate character recognition techniques for distinguishing segmented CAPTCHA characters.

\subsection{Dark Web Data Collection for CTI}
Cyber attacks are projected to cost the global economy \$6 trillion by 2021 \cite{morgan_2017_2017}. The mitigation of cyber attacks increasingly relies on gathering hacker intelligence from the dark web \cite{chen_dark_2012} to proactively gain reconnaissance on emerging cyber threats and key threat actors \cite{ebrahimi_semi-supervised_2020}. The dark web features a conglomerate of covert illegal platforms, including shops selling stolen credit/debit card and dark net marketplaces facilitating transactions between cybercriminals. Accordingly, dark web data collection has been considered as a key step in developing proactive CTI  \cite{Chiang10.1145/2361256.2361257, robertson_darkweb_2017}. 

Automated data collection from the dark web is crucial to proactively responding to cyber threats and data breach incidents. However, the automation of dark web data collection is complicated by anti-crawling measures widely adopted in the dark web \cite{samtani2017exploring}. These measures include user authentication, session timeout, deployment of cookies, and CAPTCHA. While most of these measures can be effectively circumvented through implementing automated counter measures in a crawler program, CAPTCHA is the most hampering anti-crawling measure in the dark web that cannot be easily circumvented due to high cognitive capabilities that are often not possessed by automation tools \cite{du_identifying_2018, zhang2019survey}. There are four major types of CAPTCHA: text-based, image-based, video-based, and audio-based. Text-Based CAPTCHA requires subjects to recognize the characters shown in a deliberately obfuscated alphanumeric pattern. Image-based CAPTCHA asks subjects to perform certain actions (e.g., drag and drop) on a specific portion of a given image. Video-based CAPTCHA challenges subjects to choose an option that best describes the content of a given video. Finally, audio-based CAPTCHA plays an audio and requires users to enter the characters mentioned in the audio. Text-based CAPTCHA is the most prevalent type among various types of CAPTCHA in the dark web, and is the focus of our research \cite{wu2019machine}. We note that despite the ubiquity of traditional text-based CAPTCHA in online space, dark web CAPTCHA patterns are uniquely positioned due to the use of more complicated background noise. Accordingly, while this study is mainly focused on dark-web CAPTCHA as a more challenging problem, the proposed method in this study is expected to be applicable to other types of CAPTCHA without loss of generality.

\subsection{Text-based CAPTCHA Breaking Methods }
\begin{figure}[htbp]
\centering
\includegraphics[width=0.4\linewidth]{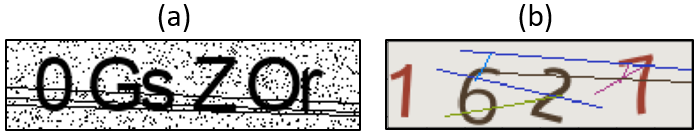}
\centering
\caption{Examples of CAPTCHA with Background and Foreground Security Measures: (a) dots+curves+font size noise (b) curves + foreground noise}
\label{captcha_example}
\end{figure}

Automated breaking of text-based CAPTCHA is non-trivial due to two main  challenges: complex security measures and variable character length. The former involves intentionally obfuscating patterns added to the CAPTCHA image for complicating the recognition task \cite{chen2017survey}. The latter draws upon the fact that most automated methods trained to break CAPTCHA with a specific length are poorly generalizable to CAPTCHAs with different character length \cite{ye2018yet}. There are two categories of security measures: foreground security measures and background security measures. Foreground security measures are mainly employed to prevent characters from having a uniform appearance. These measures include font change (i.e., varying the typeface of the character to non-standard unseen fonts), character rotation (i.e., varying the angular position of each character), and color change (i.e., varying the foreground color of each character). Background security measures provide additional obfuscation to the background of the CAPTCHA image by introducing dot noise (i.e., dot-shaped noise applied to the background surface while interfering with the characters), curve noise (i.e., irregular curvatures that cross characters), and color change (i.e., varying the color of noise so that it is not simply removable by elimination of a specific color). Figure \ref{captcha_example} illustrates examples of background and foreground security measures. 

Extant CAPTCHA-breaking frameworks often employ three steps to address these security measures \cite{wu2019machine}:
    \begin{itemize}
    \item Image Preprocessing: Performs a series of computer vision processing techniques to remove background and foreground noise
    \item Character Segmentation: Separates characters in the CAPTCA image for character-level CAPTCHA breaking (necessary to handle patterns with variable character length)
    \item Character Recognition: Identifies foreground characters via deep learning-based classification architectures such as CNN.
    \end{itemize}
We summarize prior CAPTCHA-breaking studies with focus on the methods employed at each of these steps in Table \ref{lit_overview}.

\begin{sidewaystable}
\centering
\vspace{125mm}
\caption{Selected Major ML-based CAPTCHA Breaking Studies}
\begin{center}
\resizebox*{1.0\textwidth}{!}{
\begin{tabular}{
|P{0.072\textwidth}
|P{0.23\textwidth}
|P{0.2\textwidth}
|P{0.095\textwidth}
|P{0.095\textwidth}
|P{0.13\textwidth}
|P{0.13\textwidth}
|P{0.1\textwidth}|}
    
\hline
\multirow{2}{*}{\textbf{Author}}&\multirow{2}{*}{\textbf{Focus}}&\multirow{2}{*}{\textbf{Testbed}}&\multicolumn{2}{|c|}{\textbf{Targeted Security Measure}}& \multirow{2}{*}{\shortstack{\textbf{Image Processing}}}& \multirow{2}{*}{\shortstack{\textbf{Character}\\\textbf{ Segmentation}}}& \multirow{2}{*}{\shortstack{\textbf{Character}\\ \textbf{Recognition}}} \\
\cline{4-5}
 & & &\textbf{Foreground} & \textbf{Background} & & &\\
\hline

\cline{1-8} 
Nouri and Rezaei\cite{nouri2020deep}&Utilize CNN to break synthesized CAPTCHA for vulnerability study&500,000 randomly synthesized CAPTCHA&Font, rotation, color&Noise, curve , color&Grayscale Conversion, 
Gaussian Smoothing&No&CNN\\
\hline
\cline{1-8}
Wu et al.\cite{wu2019machine}&CAPTCHA breaking for numerous Chinese characters&Chinese National Enterprise Credit Information Publicity System (10k)&Font, rotation, color&Noise,color&Grayscale Conversion, Gaussian Smoothing&Yes&CNN\\
\hline
\cline{1-8}
Ferreira et al.\cite{ferreira2019breaking}&Apply sparsity constraint in CNN to improve accuracy&Kaggle CAPTCHA dataset (87,047 images)&Font , rotation&-&Dilation, Erosion&Yes&CNN\\
\hline
\cline{1-8}
Ye et al.\cite{ye2018yet}&Use generative adversarial network to remove security features&33 Captcha sets (MS, eBay, Sina ,etc.)&Font, rotation,  color&Noise, curve , color&GAN-Based Background Removing&No&CNN\\
\hline
\cline{1-8}
Tang et al.\cite{tang2018research}&Leverage segmentation and deep learning to break English and Chinese CAPTCHA &11 Captcha sets&Font, rotation, color&Color&Grayscale Conversion,Gaussian Smoothing&Yes&CNN\\
\hline
\cline{1-8}
George et al.\cite{george2017generative}&Design a structured probabilistic graphical model for CAPTCHA breaking &4 surface web CAPTCHA sets (reCAPTCHA, Botdetect, Yahoo etc.)&Font, rotation,  color&Noise, curve, color&-&Yes&RCN\\
\hline
\cline{1-8}
Le et al.\cite{le2017using}&Utilize synthetic data to train a deep learning model to break CAPTCHA&7 platforms (Baidu, eBay, Yahoo)&Font, rotation&-&-&No&CNN and RNN\\
\hline
\cline{1-8}
Hussain et al.\cite{hussain2017segmentation}&Leverage segmentation and deep learning to break the CAPTCHA with  overlapping characters&Taobao, eBay, and MSN (3,500 CAPTCHA images)
&Font, rotation&-&Normalization, Grayscale Conversion&Yes&ANN\\
\hline
\cline{1-8}
Gao et al.\cite{gao2013robustness}&Design a CNN-based solution to  break hollow CAPTCHA by applying CFS segmentation method .&Yahoo (1,500 CAPTCHA images)
&Font, rotation&Curve, color&Normalization&Yes&ANN\\
\hline

\end{tabular}
\renewcommand{\arraystretch}{2}
}
{\raggedright Note:  ANN: Artificial Neural Network; CNN :  Convolutional Neural Network; GAN: Generative Adversarial Network; CFS: Color Filling Segmentation; RNN: Recurrent Neural Network; RCN: Recursive Cortical Network.\par}
\label{lit_overview}
\end{center}
\end{sidewaystable}

\subsubsection{Image Processing and Background Denoising}
Preprocessing CAPTCHA images involves applying computer vision techniques to enhance the foreground and denoise the background. Past studies utilize five major image preprocessing techniques for this purpose: normalization, grayscale conversion, Gaussian smoothing, dilation, and erosion. Normalization scales pixel values into a certain range to enhance inconspicuous features \cite{shanmugamani2018deep}. Grayscale conversion changes colored images into grayscale to reduce the negative impact of a high variance in colors \cite{gao2013robustness}. Gaussian smoothing applies Gaussian function to remove details and reduce the impact of curves and dots on the detection of characters’ edges \cite{wu2019machine}. Dilation enlarges shapes proportionally to help repair the missing parts in characters \cite{ferreira2019breaking}. Finally, erosion down-scales shapes proportionally in order to shrink dot noises \cite{ferreira2019breaking}. Table \ref{img_processing} summarizes the main strengths of each image preprocessing technique in dealing with background noise. While these techniques help enhance foreground characters, background denoising in complex patterns still remains a challenge \cite{gao2013robustness}. As shown in Table 2, the removal of curve noise is a difficult task that is hard to address with common image preprocessing methods. This is mainly due to the resemblance of the curves to the foreground characters. Accordingly, attempts to remove curves may result in unintended removal of foreground characters. Background denoising is even more difficult for dark web CAPTCHA with complex and noisy backgrounds. 

\begin{table*}
\centering
\caption{Image processing techniques to address background security features in CAPTCHA   
}
\begin{center}
\resizebox*{0.95\textwidth}{!}{
\begin{tabular}{
|P{0.2\textwidth}
|P{0.35\textwidth}
|P{0.1\textwidth}
|P{0.1\textwidth}
|P{0.1\textwidth}|}

\hline
\multirow{2}{*}{\textbf{Method}}&\multirow{2}{*}{\textbf{Major Goal}}&\multicolumn{3}{|c|}{\shortstack{\textbf{Targeted Background} \\ \textbf{Security Measure}}} \\
\cline{3-5} 
&&\textbf{Dot Noise} & \textbf{Curve Noise} &\textbf{Color}\\
\hline

\cline{1-5} 
Normalization \cite{hussain2017segmentation}&Enhance inconspicuous features in images.&\ding{53}&\ding{53}&\checkmark\\
\hline
\cline{1-5} 
Grayscale Conversion \cite{tang2018research}&Omit the impact of different colors.&\ding{53}&\ding{53}&\checkmark\\
\hline
\cline{1-5} 
Gaussian Smoothing \cite{wu2019machine}&Lower the impact of dot noise on the character edges.&\checkmark&\ding{53}&\ding{53}\\
\hline
\cline{1-5} 
Dilation \cite{shanmugamani2018deep}&Fix the missing parts of characters.&\ding{53}&\ding{53}&\ding{53}\\
\hline
\cline{1-5} 
Erosion \cite{shanmugamani2018deep}&Shrink the dots’ size as much as possible.&\checkmark&\ding{53}&\ding{53}\\
\hline

\end{tabular}
}
{\raggedright Note:  ANN: Artificial Neural Network; CNN :  Convolutional Neural Network; GAN: Generative Adversarial Network; CFS: Color Filling Segmentation; RNN: Recurrent Neural Network; RCN: Recursive Cortical Network.\par}
\label{img_processing}
\end{center}
\end{table*}                                                                          

While deep learning models have demonstrated promising results in image processing and can potentially improve background removal, most deep learning models are required to be trained on a large number of unseen background patterns. However, there lacks enough training data of background patterns for training such models. This scarcity is even more severe for specific domains such as the dark web. Recently, Ye et al. \cite{ye2018yet} have shown that Generative Adversarial Network (GAN) can address this issue by automatically generating background patterns with eliminated curve noise. GAN for background removal comprises two competing neural networks known as generator and discriminator. The generator tries to generate patterns with clean background from input CAPTCHAs. The discriminator tries to identify if the generator has fully removed the background. Nevertheless, the approach in \cite{ye2018yet} operates at the image level, and thus is not applicable to dark web CAPTCHA with variable length.

\subsubsection{Character Segmentation to Address Variable Character Length}
Despite the noisy background, another unique challenge of dark web CAPTCHA is the variable character length. Many automated CAPTCHA breaking solutions operate at the image level, where the CAPTCHA image is analyzed in its entirety and the CAPTCHA breaker predicts characters in the CAPTCHA altogether. As such, these solutions need to be trained on fixed-length CAPTCHA. The variable character length makes such image-level CAPTCHA breakers ineffective for two major reasons. First, image-level models that are trained on a pre-specified length are not applicable to CAPTCHAs with a different length of characters. Second, at the image level, the number of class labels grows exponentially with the number of characters (e.g., $10^3$ for 3-digit and $10^4$ for 4-digit numerical CAPTCHA). As such, a very large number of CAPTCHA images is required for model training because each class label needs to have a sufficient number of training CAPTCHA images \cite{ye2018yet}.

In contrast, character-level CAPTCHA breakers require a much smaller number of class labels (e.g., 10 different classes (i.e., {0,…, 9}) for numerical patterns). In particular, character-level CAPTCHA breakers perform character segmentation to separate characters from other characters prior to recognition \cite{wu2019machine}. Four segmentation methods are commonly leveraged in CAPTCHA breaking research: color filling segmentation (CFS), interval-based, pixel distribution-based, and contour detection-based. We provide a review of these methods in Table \ref{seg_techniques}. As observed in the table, the contour detection method is more suitable for the dark web CAPTCHA since the method can operate despite the changes in font, color, and rotation of the characters. This is mainly due to the independence of the contours from these security measures.

\begin{table*}
\centering
\caption{Character segmentation methods for CAPTCHA breaking}

\begin{center}
\resizebox*{1\textwidth}{!}{
\begin{tabular}{
|P{0.15\textwidth}
|P{0.3\textwidth}
|P{0.15\textwidth}
|P{0.125\textwidth}
|P{0.15\textwidth}
|P{0.125\textwidth}|}

\hline
\multirow{2}{*}{\textbf{Method}}&\multirow{2}{*}{\textbf{Major Goal}}&\multirow{2}{*}{\textbf{Source}}&\multicolumn{3}{|c|}{\textbf{Targeted Background Security Measure}} \\
\cline{4-6} 
&&&\textbf{Dot Noise} & \textbf{Curve Noise} &\textbf{Color}\\
\hline

\cline{1-6} 
CFS&Fills hollow shapes with a different color to differentiate characters.&Gao et al., 2013 \cite{gao2013robustness}&\checkmark&\ding{53}&\checkmark\\
\hline
\cline{1-6} 
Interval-based Segmentation&Split CAPTCHA images by fixed intervals.&Tang et al., 2018 \cite{tang2018research}&\ding{53}&\checkmark&\ding{53}\\
\hline
\cline{1-6} 
Pixel Distribution&Crops the characters based on the variance of the pixel values.&Hussain et al., 2016 \cite{hussain2017segmentation}&\checkmark&\ding{53}&\ding{53}\\
\hline
\cline{1-6} 
Contour Detection&Identifies an area containing one single character based on the contour features of the character.&Ferreira et al., 2019 \cite{ferreira2019breaking}&\checkmark&\checkmark&\checkmark\\
\hline

\end{tabular}
}
{\raggedright Note:  ANN: Artificial Neural Network; CNN :  Convolutional Neural Network; GAN: Generative Adversarial Network; CFS: Color Filling Segmentation; RNN: Recurrent Neural Network; RCN: Recursive Cortical Network.\par}
\label{seg_techniques}
\end{center}
\end{table*}
\subsubsection{Character Recognition}
After identifying the character boundaries via segmentation, the last step involves correctly recognizing the characters within each boundary \cite{gao2013robustness}. As shown in Table \ref{lit_overview}, among the past studies, Convolutional Neural Networks (CNNs) have been widely used for character recognition task \cite{weng2019towards,chen2017survey}. CNNs have shown promising results in counteracting foreground security measures such as rotation, color change, and font change of the characters. A CNN is built upon three main components: convolution layer, sampling layer, and fully connected layer. Each CNN component contributes to a certain aspect of CAPTCHA Character recognition. Convolution layer extracts geometrical salient features from local regions of the input image. For CAPTCHA breaking, this layer extracts rotation-invariant features from characters. The sampling layer combines features from local regions to generate more abstract features. For CAPTCHA breaking, this layer helps identify characters despite differences in their font and sizes. The fully connected layer weights the extracted features and assigns a probability to the output. For CAPTCHA breaking, this layer predicts the pattern’s class label based on the extracted features.
Accordingly, we expect that CNN can effectively contribute to dark web CAPTCHA character recognition after proper character segmentation.
\subsection{Research Gaps and Questions}
Existing CAPTCHA breaking methods are not designed for addressing specific characteristics of the dark web CAPTCHA. As such, security analysts often face trouble with effectively facilitating the automated collection of dark web content for CTI. Specifically, two major research gaps are identified from the review of prior studies. First, there lack studies offering methods for breaking CAPTCHA with complex noisy backgrounds in the dark web. Second, existing methods do not address CAPTCHA with variable character lengths and noisy backgrounds within a unified framework.
Based on these gaps, we pose the following research question:
\textit{How can we design an automated CAPTCHA breaking framework to address the noisy background and variable character length in the dark web?} 
\section{Research Design}
\begin{figure}[htbp]
\centering
\includegraphics[width=0.4\linewidth]{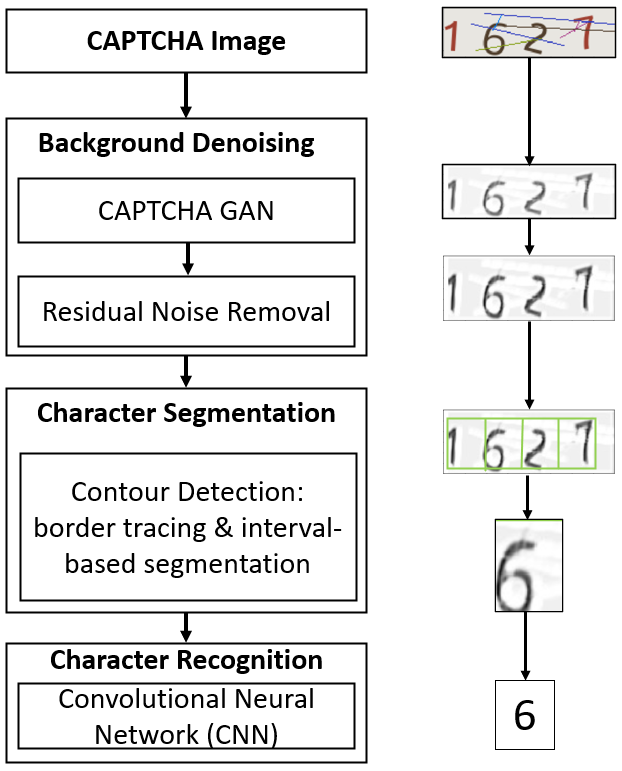}
\centering
\caption{DW-GAN Framework for Breaking Dark Web-specific CAPTCHA Images (left) and Corresponding Illustration (right) }
\label{fig_framework}
\end{figure}

To address our research question, we propose Dark Web-GAN (DW-GAN), a novel GAN-based framework that utilizes background denoising, character segmentation, and character recognition to automatically break dark web CAPTCHA. DW-GAN was implemented using PyTorch and OpenCV open-source libraries. DW-GAN aims to counteract background security measures and address the challenge of variable character length for dark web CAPTCHA. Consistent with the steps described in the literature review, DW-GAN comprises three major components shown in Figure \ref{fig_framework}.

As illustrated in the figure, first, the GAN-based background denoising component removes the noisy background from a given dark web CAPTCHA image. Then, the denoised CAPTCHA is processed using Gaussian smoothing for removing the residual noise. As opposed to the GAN approach used by Ye et al. {\cite{ye2018yet}} that operates at the image level, and thus is not applicable to dark web CAPTCHA with variable length, our proposed DW-GAN is designed to recognize CAPTCHA at the character level, which does not depend on the CAPTCHA length. Once the background is denoised, characters are segmented using an enhanced contour detection algorithm so that the remaining steps will not depend on the character length of the CAPTCHA image. To this end, we enhance the core border tracing algorithm with a subsequent region enlargement procedure, which is described later. Lastly, segments of the original CAPTCHA image are fed to a CNN for deep learning-based character recognition. We detail each component of DW-GAN in the following subsections.

\subsection{Background Denoising}
\subsubsection{Background Denoising: CAPTCHA GAN}
Attaining a training dataset encompassing all possible background patterns might not be practical for dark web CAPTCHA as it requires expensive human-labeled data and excessive human involvement. As such, we propose to leverage GAN to automatically generate CAPTCHA background patterns for training background denoising. Specifically, the background denoising component in DW-GAN consists of two sub-components: (1) the CAPTCHA GAN sub-component, a customized GAN architecture to automatically learn how to remove background noise, and (2) the residual background noise removal sub-component to further enhance the CAPTCHA foreground. CAPTCHA GAN aims to reduce various background curve noises in CAPTCHA images automatically. This is achieved by incentivizing the CAPTCHA GAN generator to generate background noise-free counterparts of original CAPTCHA images. This generative feature of CAPTCHA GAN allows for training the model with only a small labeled dataset (i.e., ~500 dark web CAPTCHA images) and achieving a high performance in background denoising. The learning process for background removal in CAPTCHA GAN includes four major steps:
    
    \begin{itemize}
        \item Step 1: The generator seeks to create background noise-free CAPTCHA image $x$ from the original dark web CAPTCHA image $y$.
        \item Step 2: The generated pattern and the corresponding original CAPTCHA image are fed to the discriminator to assess whether the background noise has been completely removed.
        \item Step 3: The generator improves by learning via minimizing the loss function $\mathcal{L_G}$, a pixel-to-pixel cross-entropy loss that compares the generator's output and the noise-free CAPTCHA image $x$. As in the conventional GAN {\cite{goodfellow2016deep}}, the loss minimization is conducted via gradient descent as part of the back propagation in the generator's neural network parameterized by network weights $\theta_g$. The generator's loss function is given by $\mathcal{L_G} = \nabla_{\theta_g}\frac{1}{N}\sum_n^N[\log(1-D(G(y_n)))]$.
        \item Step 4: The discriminator improves by maximizing the loss function $\mathcal{L_D}$ that compares the true label (i.e., incomplete (0) vs. complete background removal (1)) and the discriminators' output. the loss minimization is conducted via gradient ascent as part of the back propagation in the discriminator's neural network parameterized by network weights $\theta_d$. The discriminators' loss is given by $\mathcal{L_D} = \nabla_{\theta_d}\frac{1}{N}\sum_n^N[(\log(D(x_n))) + \log(1-D(G(y_n)))]$. to make the discrimination more accurate even on the denoised CAPTCHA image from generator's output.
    \end{itemize}

\begin{figure}[htbp]
\centering
\setlength{\abovecaptionskip}{0cm}

\includegraphics[width=0.45\textwidth]{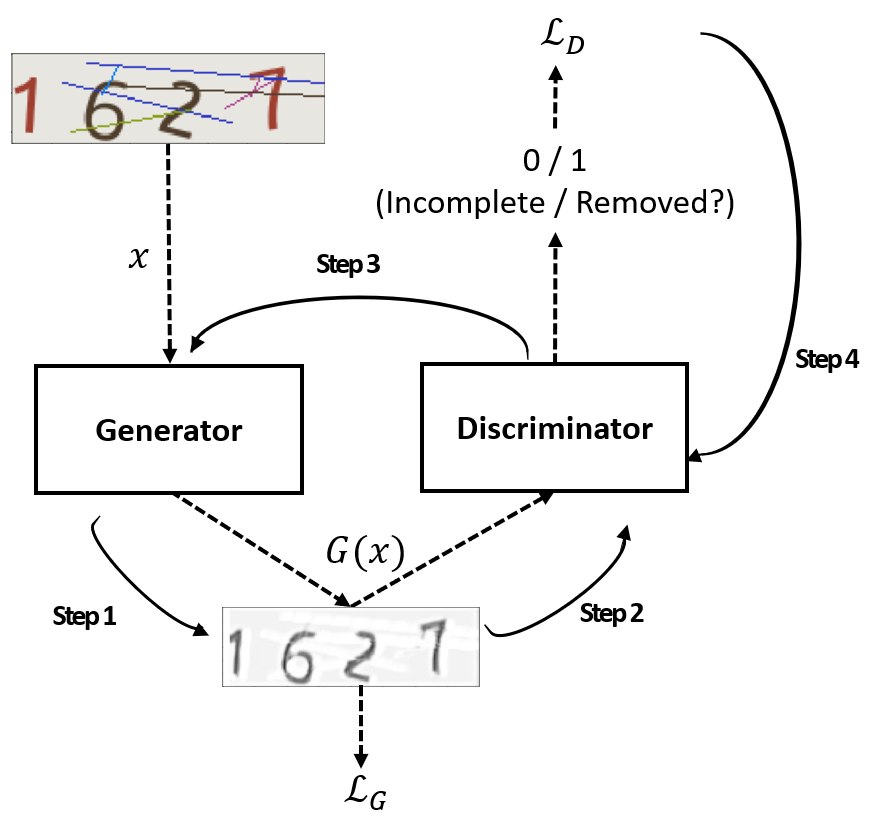}
\caption{Abstract View of CAPTCHA GAN for Background Denoising}
\setlength{\belowcaptionskip}{4cm}
\label{fig_captchagan}
\end{figure}

Steps 1-4 are repeated until the generator and discriminator reach the equilibrium condition, where neither is able to improve the performance further \cite{goodfellow2016deep}. Figure \ref{fig_captchagan} depicts the abstract view of these steps.

CAPTCHA GAN's process in Figure {\ref{fig_captchagan}} can be formalized as the minimax game with value function $V$ in Equation \ref{main_loss}, in which $G(x_n)$ and $D(x_n)$ denote the generator and discriminator networks, respectively.

\begin{equation}
  \underset{G}{\operatorname{min}}\
  \underset{D}{\operatorname{max}}\ V
  (D, G) = \frac{1}{N}\sum_n^N[(\log(D(x_n))) + \log(1-D(G(y_n)))]
  \label {main_loss}
\end{equation}

$N$ denotes the number of images in the dataset and $x_n$ represents the input CAPTCHA image. Also, $G(x_n)$ associates with $\mathcal{L_G}$ and $D(x_n)$ corresponds to discriminator's loss $\mathcal{L_D}$. CAPTCHA GAN was implemented in Python using PyTorch deep learning library. All training and testing processes for experiments in Section \ref{eval} were executed on a single Nvidia RTX 2070 GPU with 2,560 CUDA cores and 8 GB internal memory. To enhance reproducibility, the model specifications of CAPTCHA GAN, including the number of layers, neurons, and activation functions are given in Appendix \ref{appendix}. Also, the implementation of CAPTCHA GAN is available at https://github.com/johnnyzn/DW-GAN as part of the DW-GAN's source code.

\subsubsection{Background Denoising: Residual Noise Removal}
\begin{figure}[htbp]
\includegraphics[width=0.5\linewidth]{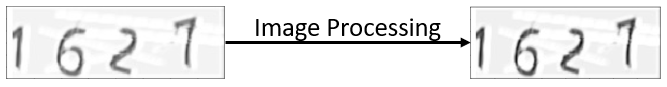}
\caption{An Example of the Result of Residual Noise Removal}
\label{fig_noiseremoval}
\end{figure}

While GAN-based background denoising seeks to remove all curve noise in the background, some residual noise may still remain. This residual noise could potentially interfere with the subsequent segmentation and character recognition. Accordingly, we utilize a combination of three image processing techniques to further remove the remaining background noise. Grayscale conversion is first leveraged to reduce the color variance as a preliminary step that alleviates color noise. Then, Gaussian smoothing is employed to reduce the visibility of the residual background noise by blurring the unnecessary details in the image. Lastly, normalization is applied to distinguish the foreground from the background color. An example of the results of residual noise removal is depicted in Figure \ref{fig_noiseremoval}.

\subsection{Character Segmentation}
\begin{figure}[htbp]
\includegraphics[width=0.3\linewidth]{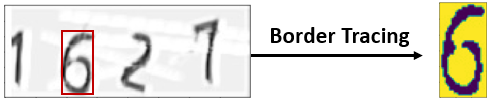}
\caption{An Example of the Result of Character Segmentation}
\label{segmentation}
\end{figure}

After background denoising, character segmentation is conducted to identify the boundary of characters. As reviewed in Table \ref{seg_techniques}, contour detection is suitable for segmenting characters in dark web CAPTCHA images due to its robustness against the dark web-specific noises. Among various contour detection methods, the \textit{border tracing} algorithm has been successfully used for segmenting CAPTCHA images since the algorithm can effectively identify the boundary pixels of the character region and separate the characters from the background \cite{yadav2018feature}. This is mainly because characters in text-based CAPTCHA often have distinguishable borders. Border tracing algorithm has two main steps \cite{pratomo2019algorithm}. In the first step, the image is converted to binary pixels (black and white) and is scanned from the upper left to the bottom right pixel. In the second step, for each pixel, the algorithm searches a square neighborhood (e.g. 3x3 pixels) to find the direction of the edges and define minimal regions to bound the character. The results of such a segmentation process is illustrated in Figure \ref{segmentation}.

\begin{figure}[htbp]
\includegraphics[width=1\linewidth]{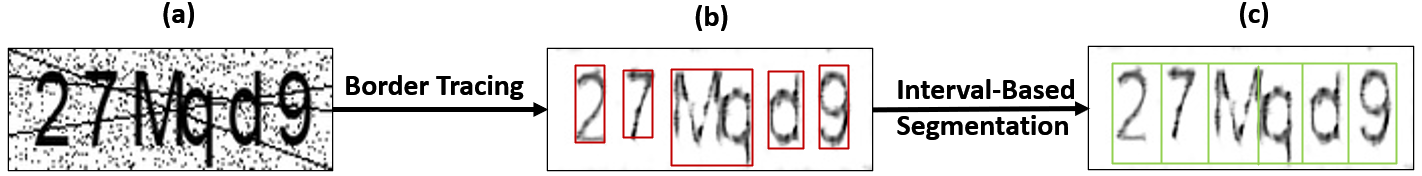}
\caption{Our Character Segmentation Process of Enhancing Border Tracing with Interval-based Segmentation}
\label{ourseg}
\end{figure}

While effective, sometimes the detected regions obtained from border tracing might not be large enough to encompass the entire character (e.g., digit ‘6’ in Figure \ref{ourseg}(b)). The core border tracing method has a tendency to yield minimal segments that may not include the entire character. As a result, incorrect segmentation would result in the false recognition of the corresponding character, and thereby the entire CAPTCHA image. To address this issue, we enhance border tracing with subsequent region enlargement to encompass the entire character via a two-step process that enhances the initial regions found by the core border tracing algorithm:

    \begin{enumerate}
        \item The initial character regions are first detected by border tracing algorithm (Figure \ref{ourseg}(b)).
        \item Subsequently, the detected regions are overlapped with the maximal regions resulted from dividing the CAPTCHA image into fixed intervals. This could result in detecting a larger region that bounds the character (Figure \ref{ourseg}(c)).
    \end{enumerate}
    
As shown in the figure, characters M and q have a considerable overlap that leads to being incorrectly identified as one character by border tracing. However, the subsequent interval-based segmentation identifies them as two imperfect characters (i.e., an incomplete "M" and a "q" with an extra stroke on the left). Given that the subsequent character recognition component is trained on various forms of incomplete characters, there is a reasonable chance that the incomplete "M" and "q" are correctly recognized. Of course, in severe cases of overlapping characters, where it could be also difficult for a human observer to distinguish characters, the character recognition component is more likely to make mistakes.

\subsection{Character Recognition}

Consistent with prior literature \cite{weng2019towards, chen2017survey}, we leverage CNN to detect characters in extracted CAPTCHA segments. Our character recognition CNN stacks convolutional layers and sampling layers to detect characters via the architecture described in Appendix \ref{appendix}. The convolutional layer extracts features from local regions of the segmented CAPTCHA image and the sampling layer combines extracted features across multiple local regions to identify fine-grained features (e.g., lines and edges). Another convolutional-sampling structure of such kind is then stacked over to extract features from larger regions through combining lines and edges into more abstract features informative of characters. Such a stacked structure further addresses CAPTCHA security measures such as rotation and font size change by jointly considering features from multiple local regions. The extracted features are then used to detect characters in a fully connected layer. Given the successful CNN training practices in the literature \cite{Pierazzi10.1145/3382158,Unger10.1145/3386243}, we used Adam optimizer \cite{kingma_adam:_2015} to minimize the cross entropy loss to train the model and utilize Rectified Linear Units (ReLU) activation function \cite{nair2010rectified} and the dropout mechanism to improve the efficiency of model training.

\section{Evaluation}
\label{eval}
We systematically evaluated the performance of our proposed framework in braking CAPTCHA images with a testbed encompassing dark web CAPTCHA datasets. In particular, our CAPTCHA testbed comprised text-based CAPTCHA images from three different dark web datasets and a widely used open-source CAPTCHA synthesizer. The dark web datasets included three sets of CAPTCHA images: two sets from carding shops (Rescator-1 and Rescator-2) and one set from a newly-emerged dark net market (Yellow Brick), as showing in Table \ref{testbed}. All three datasets were suggested by CTI experts. These datasets were selected based on their popularity and scale in the dark web. For each dataset, a TOR-routed spider was developed to collect 500 CAPTCHA images. The three datasets were labeled and inspected by two CTI experts. The second data source of our testbed was an open-source CAPTCHA synthesizer. We leveraged the synthesizer to generate CAPTCHA images with controllable security measures that enabled us to create a fair comparison for different methods on CAPTCHA images with variable character length. These datasets along with their CAPTCHA examples are summarized in Table \ref{testbed}. 
\begin{table}[h]
\centering
\caption{Summary of our dark web CAPTCHA testbed}
\begin{center}
\begin{tabular}{
|m{1.6cm}<{\centering}
|m{2.6cm}<{\centering}
|m{1.2cm}<{\centering}
|m{1.7cm}<{\centering}
|m{1.6cm}<{\centering}
|m{3.5cm}<{\centering}
|}
\hline
\textbf{Category} &\textbf{Source} &\textbf{Amount} &\textbf{Character Length} &\textbf{Character Type}&\textbf{Sample}\\
\hline
\multirow{3}{*}&\textbf{Rescator-1}&500&4&Digit&\vspace{0.9mm}\includegraphics[width=0.6\linewidth]{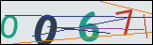}\\
\cline{2-6}
\textbf{Dark Web CAPTCHA}&\textbf{Rescator-2}&500&5&Digit&\vspace{0.9mm}\includegraphics[width=0.6\linewidth]{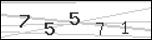}\\
\cline{2-6}
&\textbf{Yellow Brick}&500&6&Digit+Letter&\vspace{1.5mm}\includegraphics[width=0.6\linewidth]{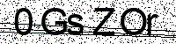}\\
\hline
\multicolumn{2}{|c|}{\textbf{Open-Source Synthesizer}}&11,000 for each character length&4, 5, 6, 7&Digit+Letter&\includegraphics[width=0.6\linewidth]{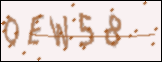}\\
\hline

\end{tabular}
\label{testbed}
\end{center}
\end{table}

We conducted three experiments to evaluate the effectiveness of our proposed research design:
\begin{itemize}
    \item Experiment 1 examined the overall CAPTCHA breaking performance of our proposed framework in comparison with the state-of-the-art baseline methods. This experiment targets the dark web-specific condition of lacking labeled CAPTCHA images for training. To this end, we sampled 500 images from each of the dark web CAPTCHA datasets serving as the evaluation testbed. The CAPATCHA breaking performance was measured by success rate \cite{bursztein2011text,ye2018yet}, a commonly used metric representing the percentage of CAPTCHA images that are successfully recognized \cite{ye2018yet}. A successful recognition entails correctly recognizing \textit{all} characters in the CAPTCHA image. Three state-of-the-art methods from past research, were used as benchmarks: Image-level CNN \cite{le2017using}, Image-level CNN with preprocessing (grayscale conversion, normalization, and gaussian smoothing) \cite{nouri2020deep}, and Character-level CNN with interval-based segmentation \cite{tang2018research}.
    
    \item Experiment 2 investigated the effect of each component in our proposed DW-GAN framework through an ablation analysis. To this end, the framework's major components, including background denoising, border tracing segmentation, and interval-based segmentation were removed from DW-GAN framework to evaluate their impact on the automated CAPTCHA breaking performance. The ablation analysis were thoroughly conducted on all dark web datasets used in benchmark evaluation in Experiment 1. Also, consistent with Experiment 1, the success rate was adopted as the evaluation criterion.
    
    \item Experiment 3 evaluated the efficacy of DW-GAN components to determine whether background denoising and character segmentation improved CAPTCHA breaking performance. Accordingly, this experiment encompassed two sub-experiments as detailed below. We employed the CAPTCHA synthesizer to generate a wide range of background noise with variable lengths. 
    \begin{itemize}
        \item Experiment 3.1 examined the contribution of the background denoising component to the overall performance in DW-GAN. We evaluated the effect of our background denoising component before and after its application to each type of background noise. To measure the effect, we adopted the structural similarity (SSIM) metirc \cite{powell2014fgcaptcha}, a widely used metric to measure the similarity of two images. SSIM can be used to determine the extent to which the background security measures are counteracted through our background denoising component. A higher SSIM between the denoised image and its clean version (without background noise) indicates a more effective removal of background security measures.
        \item Experiment 3.2 investigated the effect of the character segmentation component on the overall performance. This experiment compared the performance of DW-GAN with a benchmark image-level method and two character-level methods to evaluate how character segmentation improved CAPTCHA breaking performance when character length varied. In accordance with the common dark web CAPTCHA length, the character lengths ranged from 4 to 7. The evaluation was based on a 90\%-10\% train-test split. That is, our training set contained 50,000 CAPTCHA images (10,000 images for each character length) and the testing set included 5,000 CAPTCHA images (1,000 images for each character length).
    \end{itemize}
\end{itemize}

\subsection{Results of Experiment 1: Benchmark Evaluation}
We compared the performance of DW-GAN to the state-of-the-art CAPTCHA breaking methods across the dark web datasets. The success rates for each method are shown in Table \ref{ex1_result}. The asterisks denote the statistical significance obtained from paired t-test between the results of DW-GAN and the above-mentioned benchmark methods (P-values are significant at 0.05:*, 0.01:**, 0.001:***).

\begin{table}[!ht]
\vspace{-2mm}
\centering
\caption{Benchmark Evaluation of DW-GAN against the state-of-the-art automated CAPTCHA breaking methods across different CAPTCHA datasets from the dark web}
\begin{center}
\vspace{-4mm}
\begin{tabular}{
|P{3.5cm}
|P{1.5cm}
|P{1.5cm}
|P{1.5cm}
|}
\hline
\multirow[b]{2}{*}{\textbf{Method}}&\multicolumn{3}{|c|}{\textbf{Dark Web CAPTCHA}} \\
\cline{2-4} 
&\textbf{Rescator-1}&\textbf{Rescator-2}&\textbf{Yellow Brick} \\
\hline
Image-level CNN only \cite{le2017using}&63.57\%***&35.05\%***&5.88\%*** \\
\hline
Image-level CNN + Preprocessing \cite{nouri2020deep} &70.5\%**&36.04\%***&7.84\%*** \\
\hline
Character-level CNN + Segmentation \cite{tang2018research}&88.12\%**&77.23\%**&93.72\%* \\
\hline
DW-GAN (Ours)&\textbf{94.4\%}&\textbf{97.50\%}&\textbf{95.98\%} \\
\hline

\end{tabular}
\label{ex1_result}
\end{center}
\end{table}
\

Three notable observations can be made from the results of Experiment 1. First, comparing the automated CAPTCHA breaking method in \cite{le2017using} against \cite{nouri2020deep}, it is observed that preprocessing improved the success rate across all datasets. Second, the character-level method proposed by \cite{tang2018research} generally achieved a higher success rate compared to image-level CAPTCHA breaking methods. Third, and most importantly, our proposed framework with GAN-based background denoising yielded the highest success rate across all datasets. Overall, DW-GAN outperformed the state-of-the-art benchmark methods with statistically significant margins in dark web CAPTCHA. DW-GAN's performance is attributed to combining background denoising and character recognition in a unified framework. The contribution of each component is further investigated in Experiment 2 and Experiment 3.

\
\subsection{Results of Experiment 2: Ablation Analysis}
To gauge the contribution of each component in the proposed DW-GAN framework, we further evaluated the CAPTCHA breaking performance (i.e., success rate) by eliminating each major component in DW-GAN. To retain the basic functionality of CAPTCHA breaking, we utilized at least one segmentation method as well as the CNN for character recognition. Accordingly, we eliminated the background denoising component, border tracing segmentation, and interval-based segmentation in a consecutive manner resulting into three alternative models to compare with the proposed DW-GAN. The results of the ablation analysis for these three models across the dark web datasets in Experiment 1 are given in Table {\ref{ex_ablation}}.

\begin{table}[!h]
\vspace{-2mm}
\centering
\caption{Ablation analysis of each component in the DW-GAN's framework}
\begin{center}
\vspace{-4mm}
\begin{tabular}{
|P{2.8cm}
|P{1.5cm}
|P{1.5cm}
|P{1.5cm}
|P{3.3cm}
|}
\hline
\multirow[b]{2}{*}{\textbf{Model}}&\multicolumn{3}{|c|}{\textbf{Dark Web CAPTCHA}}& \multirow[b]{2}{*}{\textbf{Average Performance}} \\
\cline{2-4} 
&\textbf{Rescator-1}&\textbf{Rescator-2}&\textbf{Yellow Brick}& \\
\hline
DW-GAN without Background Denoising & 93.4\%&88.12\%&21.96\%&67.83\% \\
\hline
DW-GAN without Border Tracing Segmentation &77.2\%&88.91\%&38.82\%&68.21\% \\
\hline
DW-GAN without Interval-Based Segmentation &63.2\%&\textbf{98.02}\%&10.00\%&57.07\% \\
\hline
DW-GAN &\textbf{94.4\%}&97.50\%&\textbf{95.98}\%&\textbf{95.96}\% \\
\hline

\end{tabular}
\label{ex_ablation}
\end{center}
\end{table}

As shown in Table {\ref{ex_ablation}}, all three essential components of DW-GAN in our design significantly contributed to the performance of automated CAPTCHA breaking on average across all three datasets. Specifically, eliminating the background denoising component resulted in approximately 28\% performance loss (67.83\% vs. 95.6\% in Table 6) on average across the three datasets. Similarly removing border tracing segmentation and interval-based segmentation leads to a 27\% and 40\% performance drop, respectively (68.21\% and 57.07\% vs. 95.96\%). These results indicate that background denoising, border tracing segmentation, and interval-based segmentation improved the performance significantly. We note that for Rescator2, due to less complicated background, border tracing alone could be sufficient for effective automated CAPTCHA breaking, and thus there is less need for interval-based segmentation. That is, our DW-GAN framework outperformed with a larger margin on more complex backgrounds with various noise types.

\subsection{Results of Experiment 3.1: Evaluating Background Denoising}
\begin{figure}
\includegraphics[width=0.7\linewidth]{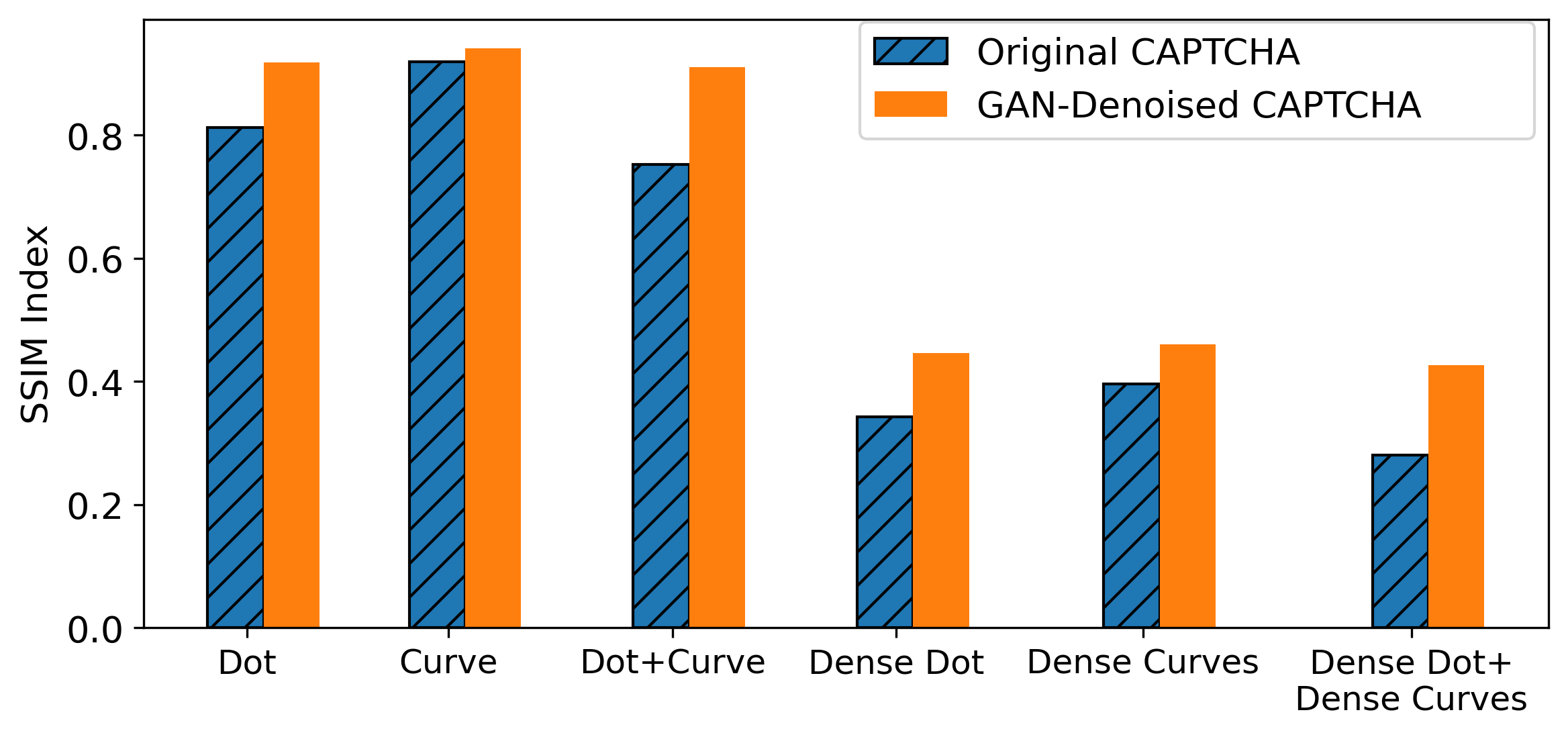}
\caption{SSIM Change Before and After Background Denoising for Different Security Measure Combinations}
\label{barchart}
\end{figure}

To evaluate the effect of our background denoising component, we first compared the SSIM of original CAPTCHA images with different types of background noise to clear CAPTCHA images (same pattern without any background noise). Then we applied our proposed GAN-based background denoising on the original image and computed the SSIM with respect to the clear CAPTCHA image. The results of these two comparisons are shown in paired bars for each background noise type in Figure \ref{barchart}. As seen in the figure, our GAN-based background denoising method shows consistently higher similarity to the clear image for various types of background noise, including dot, curves, dot+curves, dense dots, dense curves, and dense dot with dense curves.

The high SSIM index achieved by DW-GAN suggests that GAN-based background denoising is capable of removing a wide variety of background noise that are widely used in the dark web.

\subsection{Results of Experiment 3.2: Evaluating Character Segmentation}
To gauge the effect of our proposed segmentation component in DW-GAN, we applied our segmentation method to various lengths of CAPTCHA images and compared the success rate with other image-level methods as well as common segmentation methods in character-level CAPTCHA breaking. The corresponding success rates are shown in Table \ref{ex_22}.

\begin{figure}[h]
\includegraphics[width=0.6\linewidth]{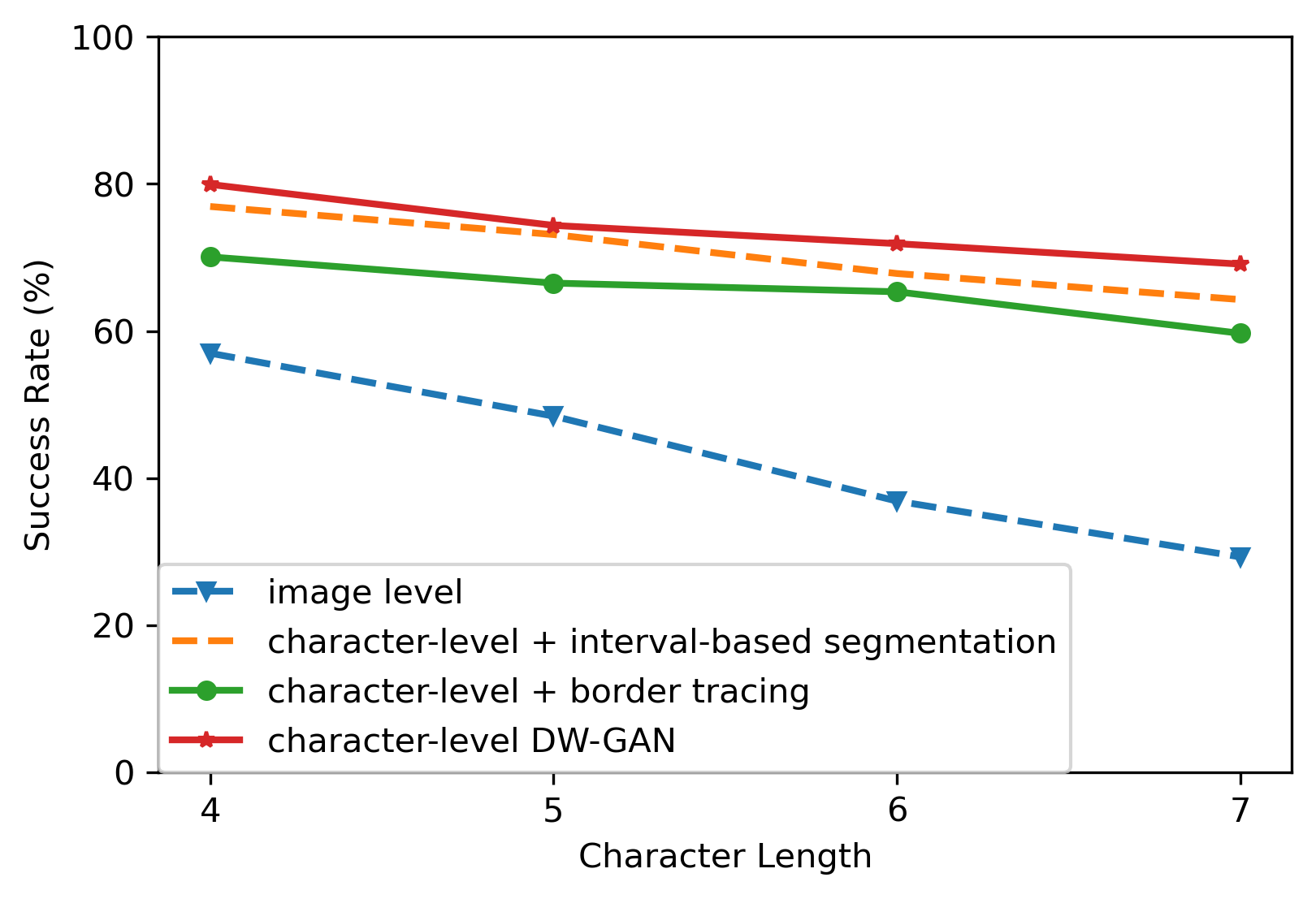}
\caption{Sensitivity of Success Rate to Increasing the Character Length}
\label{trend}
\end{figure}

\begin{table*}[!ht]
\centering
\caption{Evaluating the effect of the proposed segmentation compoent in DW-GAN on CAPTCHA patterns with variable lengths}

\begin{center}
\resizebox*{1\textwidth}{!}{
\begin{tabular}{
|P{0.1\textwidth}
|P{0.25\textwidth}
|P{0.25\textwidth}
|P{0.2\textwidth}
|P{0.2\textwidth}|}

\hline
\textbf{Character Length}&\textbf{Image-level CNN + Preprocessing}&\textbf{Character-Level + Interval-Based Seg}&\textbf{Character-Level + Border Tracing}&\textbf{DW-GAN's Segmentation} \\
\hline
4&56.98\%&76.92\%&70.07\%&79.92\% \\
\hline
5&48.43\%&73.11\%&66.51\%&74.35\% \\
\hline
6&36.88\%&67.81\%&65.33\%&71.87\% \\
\hline
7&29.29\%&64.27\%&59.68\%&69.09\% \\
\hline

\end{tabular}
}
\label{ex_22}
\end{center}
\end{table*}
We highlight two major observations from Table \ref{ex_22}. First, while the CAPTCHA breaking performance of image-level methods drastically decreased as the character length increased, the variation in character length had a significantly smaller impact on the CAPTCHA breaking performance of our proposed segmentation method in DW-GAN. This shows that DW-GAN's segmentation method performed better than other segmentation methods. Second, consistent with the result of Experiment 1, character-level methods demonstrated a higher success rate than image-level methods over all character lengths. To further demonstrate the sensitivity of success rate to increasing the character length for image-level and character-level methods, we plot the performance while increasing the character length in Figure \ref{trend}. Observing that DW-GAN maintains a high performance as character length increases signifies the effectiveness of the enhanced character segmentation proposed in DW-GAN.

\section{Dark Web Case Study: Yellow Brick DNM}

\begin{figure}[htbp]
\centering
\includegraphics[width=1\linewidth]{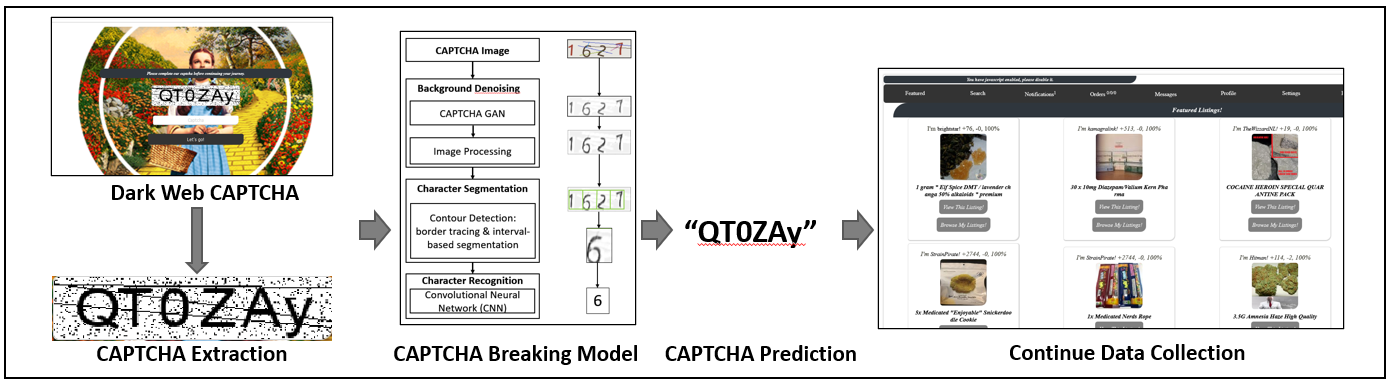}
\centering
\caption{ An Automated Process for Dark Web CAPTCHA Breaking in a Dark Net Market with DW-GAN}
\label{yellowbrick}
\end{figure}

We further demonstrate the application of our proposed framework on a real-world DNM. The purpose of this case study is two-fold. First, we sought to show that our proposed framework could detect and break the CAPTCHA to facilitate scalable, automated dark web data collection. Second, we aim to show that our proposed framework was able to effectively remove human involvement from dark web data collection. Figure \ref{yellowbrick} illustrates the process of employing DW-GAN for automated dark web CAPTCHA breaking in a DNM.

DW-GAN can be embedded in a dark web crawler. When a CAPTCHA challenge is encountered in a targeted dark web platform, DW-GAN is activated to break the CAPTCHA within a few attempts. The entire process can be broken down into several stages. First, the crawler navigates to the homepage of the targeted DNM. Parsing the login page, the crawler captures the CAPTCHA image and send it to our trained DW-GAN component. DW-GAN denoises, segments, and recognizes the CAPTCHA image. The recognized result would return to our crawler. Finally, the crawler would iteratively enter the predicted result until the correct prediction occurs (i.e., the platform permits access). As such, the DW-GAN enhanced crawler is guaranteed to collect illicit product information without human interaction.

Based on the above automated process in Fig \ref{yellowbrick}, a Tor-routed web crawler was developed and equipped with DW-GAN to automatically collect illegal advertised product pages from the Yellow Brick DNM. The crawler first detected the login page and automatically entered credentials. Then, the crawler collected all Onion links that contained the information of illegal items by sending HTTP requests to the platform server. Yellow Brick requires re-login and bypassing CAPTCHA for every 15 HTTP requests on average. Without our automated solution, the web crawler needs human interventions every three minutes to manually bypass the CAPTCHA challenges issued by the platform.

Using a crawler enhanced by our DW-GAN, we were able to collect 1,831 illegal products from Yellow Brick. Among these products, there were 286 cybersecurity-related items, including 102 stolen credit cards, 131 stolen accounts, 9 forged document scans, 44 hacking tools, and 1,223 drug-related products (opioid, cocaine, etc.). Overall, collecting Yellow Brick market with DW-GAN took about 5 hours without human involvement. In particular, each HTTP request took 8.8 seconds for loading a new webpage; therefore crawling 1,831 pages took 268.5 minutes. Solving the recurring CAPTCHA challenges (per each 15 HTTP requests) took our DW-GAN crawler 18.6 seconds. Overall, the proposed framework could automatically break CAPTCHA with no more than 3 attempts. Breaking all CAPTCHA images take about 76 minutes in total for all 1,831 product pages, a process that is fully automated.

\section{Conclusion and Future Directions}

Text-based CAPTCHA breaking has been a major challenge for collecting large-scale dark web data for developing proactive CTI. Two major challenges have been the particularly noisy background and the variable character length of CAPTCHA images. Leveraging GAN and CNN, we propose a novel and automated framework for breaking text-based CAPTCHA in the dark web. The proposed framework utilizes GAN to counteract background security measures for dark web-specific CAPTCHA and leverages an enhanced border tracing algorithm to extract segments encompassing single characters from CAPTCHA images. Our DW-GAN was evaluated on several dark web CAPTCHA datasets. DW-GAN significantly outperformed the state-of-the-art benchmark methods on all datasets in the dark web research context, where there is a lack of labeled CAPTCHA data. Moreover, the proposed framework maintained relatively consistent CAPTCHA breaking performance when character length varies. While collecting dark web data traditionally requires significant human effort, a web crawler equipped with DW-GAN is able to  collect large-scale dark web data without human involvement.

Future research is needed to counteract more sophisticated CAPTCHA patterns in the dark web. For instance, we have observed some newly emerged small-sized dark web platforms that complement the text-based CAPTCHA with a ‘question-answering’ challenge. In addition to solving the text-based CAPTCHA, these platforms require user to answer a question. For example, the user may be asked to answer queries such as "adding two numbers" or "the first month of the year." Enhancing DW-GAN for other emerging CAPTCHA challenges remains as a promising future research direction.

\begin{acks}
This material is based upon work supported by the National Science Foundation (NSF) under the following grants: SaTC-1936370, CICI-1917117, and SFS-1921485.
\end{acks}

\bibliographystyle{ACM-Reference-Format}
\bibliography{main}
    
\appendix

\section{appendix}
\label{appendix}

This section details the specifications and parameter/hyperparameter settings of DW-GAN. The architecture of DW-GAN was aimed to be parsimonious. To train the deep learning models in DW-GAN, including the generator and discriminator's networks, after a light-weight parameter tuning in a grid search manner, we adopted the learning rate of 0.0035. Also, to utilize the maximum GPU memory, we set the batch size for training process to 200. The GAN-based background denoising was effectively trained after 1,000 epochs. Given that the complexity of character recognition is often less than that of background denoising, effective training the CNN model took 100 epochs. Table {\ref{Hyper_LR}} includes selected learning rates from the parameter tuning process. 
\begin{table}[H]
\centering
\caption{Different learning rates' impact on DW-GAN success rate}

\begin{center}
\resizebox*{0.8\textwidth}{!}{
\begin{tabular}{
|P{0.2\textwidth}
|P{0.15\textwidth}
|P{0.15\textwidth}
|P{0.15\textwidth}|}
\hline
    \textbf{Learning Rate}&\textbf{0.035}&\textbf{0.0035}&\textbf{0.001}\\
\hline
DW-GAN& 59.2\% &\textbf{94.4}\%&86.6\%\\
\hline
\end{tabular}
}
\label{Hyper_LR}
\end{center}
\vspace{-5mm}
\end{table}

Table \ref{GAN_G} summarizes the generator's architecture in our CAPTCHA GAN for  background denoising. The generator has 5 convolutional layers with 64, 128, 128, 64, and 1 neural units. As noted, the filter size in convolutional layers was $3\times3$. 

\begin{table}[H]
\centering
\caption{Generator's parameter specifications for background denoising in CAPTCHA GAN}

\begin{center}
\resizebox*{0.8\textwidth}{!}{
\begin{tabular}{
|P{0.1\textwidth}
|P{0.23\textwidth}
|P{0.37\textwidth}
|P{0.15\textwidth}|}
\hline
\textbf{Layer Number}&\textbf{Element}&\textbf{Parameter}&\textbf{Value}\\
\hline
1& Convolutional Filter &[filter number, size, Stride, pad ]&[64, 3x3, 1, 1]\\
\hline
2& Convolutional Filter &[filter number, size, Stride, pad ]&[128, 3x3, 1, 1]\\
\hline
3& Convolutional Filter &[filter number, size, Stride, pad ]&[128, 3x3, 1, 1]\\
\hline
4& Convolutional Filter &[filter number, size, Stride, pad ]&[64, 3x3, 1, 1]\\
\hline
5& Convolutional Filter &[filter number, size, Stride, pad ]&[1, 3x3, 1, 1]\\
\hline
\end{tabular}
}
\label{GAN_G}
\end{center}
\end{table}
Table \ref{GAN_D} includes the specifications of discriminator's architecture in our CAPTCHA GAN. The discriminator encompasses 6 convolutional layers followed by a fully connected layer. These convolutional layers have 16, 32, 64, 128, 256, and 256 neural units, respectively. Similar to the generator, $3\times3$ filters were used in the convolutional layers.

\begin{table}[H]
\centering
\caption{Discriminator's parameter specifications for background denoising in CAPTCHA GAN}

\begin{center}
\resizebox*{0.8\textwidth}{!}{
\begin{tabular}{
|P{0.1\textwidth}
|P{0.23\textwidth}
|P{0.37\textwidth}
|P{0.15\textwidth}|}
\hline
\textbf{Layer Number}&\textbf{Element}&\textbf{Parameter}&\textbf{Value}\\
\hline
1& Convolutional Filter &[filter number, size, Stride, pad ]&[16, 3x3, 1, 1]\\
\hline
2& Convolutional Filter &[filter number, size, Stride, pad ]&[32, 3x3, 1, 1]\\
\hline
3& Convolutional Filter &[filter number, size, Stride, pad ]&[64, 3x3, 1, 1]\\
\hline
4& Convolutional Filter &[filter number, size, Stride, pad ]&[128, 3x3, 1, 1]\\
\hline
5& Convolutional Filter &[filter number, size, Stride, pad ]&[256, 3x3, 1, 1]\\
\hline
6& Convolutional Filter &[filter number, size, Stride, pad ]&[256, 3x3, 1, 1]\\
\hline
7& Fully-Connected  &[neuron number]&[1]\\
\hline
\end{tabular}
}
\label{GAN_D}
\end{center}
\end{table}
Table \ref{CNN_parameter} includes the CNN architecture used for the character-level recognition in DW-GAN. The CNN architecture has 3 convolution layers (each followed by a max-pooling layer) and 3 fully connected layers. As noted, ReLU was used as activation function.
\begin{table}[H]
\centering
\caption{Character-Level CNN parameter specifications for character recognition}

\begin{center}
\resizebox*{0.8\textwidth}{!}{
\begin{tabular}{
|P{0.1\textwidth}
|P{0.2\textwidth}
|P{0.4\textwidth}
|P{0.15\textwidth}|}
\hline
\textbf{Layer Number}&\textbf{Element}&\textbf{Parameter}&\textbf{Value}\\
\hline
\multirow{2}{*}{1}& Convolutional Filter &[filter number, size, Stride, pad ]&[32, 3x3, 1, 1]\\
\cline{2-4} 
&Max Pooling&[size, pad]&[2x2, 2]\\
\hline
\multirow{2}{*}{2}& Convolutional Filter &[filter number, size, Stride, pad ]&[64, 3x3, 1, 1]\\
\cline{2-4} 
&Max Pooling&[size, pad]&[2x2, 2]\\
\hline
\multirow{2}{*}{3}& Convolutional Filter &[filter number, size, Stride, pad ]&[64   , 3x3, 1, 1]\\
\cline{2-4} 
&Max Pooling&[size, pad]&[2x2, 2]\\
\hline
4&Fully-Connected&[neuron number]&[1024]\\
\hline
5&Fully-Connected&[neuron number]&[256]\\
\hline
6&Fully-Connected&[neuron number]&[character number]\\
\hline
\end{tabular}
}
\label{CNN_parameter}
\end{center}
\end{table}
Since the activation function and the number of convolutional layers have significant effects on the performance of the CNN model, we show selected results of their parameter tuning process to optimize the character recognition. Specifically, for the activation function, we compared ReLU, Tanh, and Softmax (Table {\ref{Hyper_AF}}). ReLU performed the best for the character recognition task in DW-GAN. Also, varying the number of convolutional layers for our CNN model indicated that three convolutional layers perform the best for CAPTCHA breaking. The comparison results are shown in Table {\ref{Hyper_Layers}}. 
\begin{table}[H]
\centering
\caption{Activation functions' impact on DW-GAN's success rate}

\begin{center}
\resizebox*{0.8\textwidth}{!}{
\begin{tabular}{
|P{0.2\textwidth}
|P{0.15\textwidth}
|P{0.15\textwidth}
|P{0.15\textwidth}|}
\hline
\textbf{Activation Function}&\textbf{ReLU}&\textbf{Tanh}&\textbf{Softmax}\\
\hline
DW-GAN& 94.4\% &74.4\%&14.8\%\\
\hline
\end{tabular}
}
\label{Hyper_AF}
\end{center}
\vspace{-5mm}
\end{table}

\begin{table}[H]
\centering
    \caption{The impact of the number of convolutional layers of the character recognition component on DW-GAN's success rate}

\begin{center}
\resizebox*{0.8\textwidth}{!}{
\begin{tabular}{
|P{0.2\textwidth}
|P{0.15\textwidth}
|P{0.15\textwidth}
|P{0.15\textwidth}|}
\hline
\textbf{Convolutional Layer Number}&\textbf{2}&\textbf{3}&\textbf{4}\\
\hline
DW-GAN& 63.2\% &94.4\%&90.2\%\\
\hline
\end{tabular}
}
\label{Hyper_Layers}
\end{center}
\vspace{-5mm}
\end{table}

\section{Online Resources}
The implementation of DW-GAN and its corresponding datasets are available at\\ https://github.com/johnnyzn/DW-GAN.

\end{document}